# Belief change with noisy sensing in the situation calculus

**Jianbing Ma, Weiru Liu and Paul Miller**
School of Electronics, Electrical Engineering and Computer Science,
Queen's University Belfast, Belfast BT7 1NN, UK
{jma03,w.liu}@qub.ac.uk, p.miller@ecit.qub.ac.uk

## Abstract

Situation calculus has been applied widely in artificial intelligence to model and reason about actions and changes in dynamic systems. Since actions carried out by agents will cause constant changes of the agents' beliefs, how to manage these changes is a very important issue. Shapiro et al. [22] is one of the studies that considered this issue. However, in this framework, the problem of noisy sensing, which often presents in real-world applications, is not considered. As a consequence, noisy sensing actions in this framework will lead to an agent facing inconsistent situation and subsequently the agent cannot proceed further. In this paper, we investigate how noisy sensing actions can be handled in iterated belief change within the situation calculus formalism. We extend the framework proposed in [22] with the capability of managing noisy sensings. We demonstrate that an agent can still detect the actual situation when the ratio of noisy sensing actions vs. accurate sensing actions is limited. We prove that our framework subsumes the iterated belief change strategy in [22] when all sensing actions are accurate. Furthermore, we prove that our framework can adequately handle belief introspection, mistaken beliefs, belief revision and belief update even with noisy sensing, as done in [22] with accurate sensing actions only.

## 1 Introduction

Situation calculus, introduced by John McCarthy [14, 15], has been applied widely to model and reason about actions and changes in dynamic systems. It was reinterpreted in [17] as *basic action theories* which are comprised of a set of foundational axioms defining the space of situations, unique-name axioms for actions, action preconditions and effects axioms, and the initial situation axioms [8]. The well known frame problem is solved by a set of special action effects axioms called *successor state axioms*.

Since actions carried out by agents cause constant changes of the agents' beliefs, developing strategies of managing belief changes triggered by actions is an important issue. The problem of *iterated belief change* within the framework of situation calculus has been investigated widely, e.g., [19, 20, 22, 13]. In [22], a new framework exceeding previous approaches was proposed in which a plausibility value is attached to every situation. This way, the framework is able to deal with nested beliefs, belief introspection, mistaken beliefs, and it can also handle belief revision and belief update (two types of belief changes) together in a seamless way.

Although this framework has many distinct advantages, it suffers from some drawbacks. In this framework, a set of initial situations which an agent considers possible are given, then every time the agent performs a sensing action, situations that do not match the sensing result (e.g., a sensing result shows the light is on, while it is considered off in the situation) are discarded. A key assumption of the framework is that all sensing actions must be accurate, an assumption that is too strong in real-world scenarios. When sensing actions are not accurate, situations that perfectly match the actual situation are in fact discarded. Discarding such situations results in inconsistency and makes subsequent reasoning unproceedable.

Let us illustrate this with the following example.

**Example 1** *(adapted from [22]) Assume that the initial situation $S_0$ is $\mathsf{InR1}(S_0) \wedge \neg\mathsf{Light1}(S_0) \wedge \neg\mathsf{Light2}(S_0)$ which states that the agent is in Room 1 ($\mathsf{InR1}(S_0)$), the lights in both Room 1 and Room 2 (assume there are only two rooms) are off. However, the agent does not know the actual situation, e.g., which room it is in and whether the lights in these rooms are on/off. So it considers two possible situations $S_1$ and $S_2$ at the beginning where*

$$S_1 = \neg\mathsf{InR1}(S_1) \wedge \mathsf{Light1}(S_1) \wedge \mathsf{Light2}(S_1)$$

$$S_2 = \mathsf{InR1}(S_2) \wedge \neg\mathsf{Light1}(S_2) \wedge \neg\mathsf{Light2}(S_2)$$

*It is easy to see that $S_2$ perfectly matches the real situation*

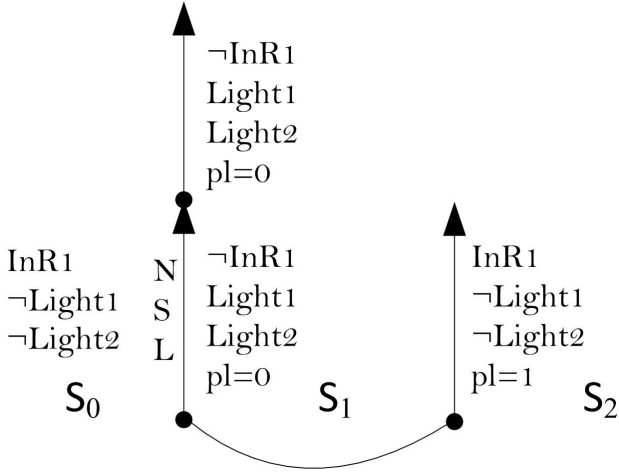

Figure 1: The correct situation is discarded when sensing is not accurate

$S_0$. *Not knowing the truth, the agent assigns $S_1$ and $S_2$ with plausibility values* 0 *and* 1 *respectively, which are the $\kappa$-rankings in [24] such that the lower the plausibility value is, the more plausible the situation is. The bottom-half of Fig.1 illustrates these three situations.*

*The agent now believes that the most plausible situation is $S_1$. To confirm, the agent performs a* SenseLight *action which senses whether the light is on in the room that the agent is currently located. A* SenseLight *(*SL*) action is* accurate *if it returns the true state of the situation and is* noisy[1] *when it returns a wrong result. In Fig. 1, a* Noisy SenseLight *action is abbreviated as* NSL. *Given that the agent is in Room 1 and the light in Room 1 is off, this sensing action is noisy which prompts the agent to believe that $S_1$ matches this sensing result whilst $S_2$ does not. As a consequence, with this framework in [22], $S_2$, the right situation, will be discarded. Now with only $S_1$ considered possible, if the agent performs another sensing action,* SenseRoom *(which senses which room the agent is located) and it is accurate, then the agent immediately find that it faces an inconsistency (that is, the result of* SenseRoom *tells the agent it is in room R1 while the only possible situation shows it is not in R1) and cannot proceed. That is, noisy sensing cannot be handled properly by the framework.*

In this paper, we extend the framework in [22] to manage noisy sensing actions. By using non-fixed plausibility values for situations, our formalism does not need to discard any situations but only changes their plausibility values, which makes it possible to accept noisy sensing actions. An important result is that we prove that rare noisy sensing does not play crucial roles in detecting the environment. More precisely, if the ratio of noisy sensing w.r.t. accurate sensing is relatively small, then an agent should be able to detect the actual situation. Furthermore, when the ratio is not that small but is restricted to a certain degree, the agent is still able to obtain a degree of chance for detecting the actual situation. On the other hand, when every sensing action is accurate, our extended framework is capable of discovering what can be derived from the framework in [22]. In addition, a set of desirable properties are proved to hold for this new formalism. In summary, we have the following main contributions.

---

[1]Note that the agent itself does not know whether a sensing action is accurate or noisy. The sensing is accurate or not is independent to the agent and the statistics of the accuracy of the sensing device (or sensing method) can be obtained in some way, such as training.

- With the new framework, an agent can continue to proceed with **noisy sensing** actions.

- We prove that it does not affect the detection of the actual situation if there are only a few noisy sensing actions. Moreover, if the ratio of noisy sensing actions is restricted to a certain degree, then the framework has a very good chance to detect the actual situation.

- When all sensing actions are accurate, the beliefs that can be induced by the framework of [22] can also be induced from our framework.

- Belief introspection, mistaken beliefs, belief revision and update can all be well handled in our framework.

The rest of the paper is organized as follows. We provide the preliminaries on situation calculus in Sec. 2. In Sec. 3, some definitions and axioms needed by our framework are introduced. In Sec. 4, properties of our framework are proposed. Sec. 5 extends Example 1 to illustrate our extended framework. Finally, we discuss some related work and conclude the paper in Sec. 6 and Sec. 7, respectively.

## 2 Background

The framework in [22] is based on an extension of the action theory [17] stemming from situation calculus [14, 15]. Here we introduce the notion of situation calculus from [22] which includes a belief operator [19, 20].

The situation calculus is a predicate calculus language for representing dynamically changing domains. A *situation* represents a snapshot of the domain. There is a set of initial situations corresponding to the ways agents believe what the domain might be initially. The actual initial state is represented by a distinguished initial situation constant, $S_0$, which may or may not be among the set of initial situations believed by an agent. The term $do(a, s)$ denotes the unique situation that results from the agent performing action $a$ in situation $s$.

A predicate (or function) whose value may change upon a situation (its last argument), is called a *fluent*. For instance,

we use the fluent $\mathsf{InR1}(s)$ to represent that the agent is in room $R1$ in situation $s$. The effects of actions on fluents are defined using successor state axioms [17], which provide a succinct representation for both effect axioms and frame axioms [14, 15]. For instance, in Example 1, if there is an action *Leave* which takes the agent from the current room to the other room, then the successor state axiom for $\mathsf{InR1}$ is:
$\mathsf{InR1}\big(do(a,s)\big) \equiv$
$\big((\neg\mathsf{InR1}(s) \land a = \mathsf{Leave}) \lor (\mathsf{InR1}(s) \land a \neq \mathsf{Leave})\big).$

This axiom says that the agent will be in Room 1 after doing action $a$ in $s$ iff either it is in Room 2 and leaves for Room 1 or is currently in Room 1 and does not leave.

Levesque [9] introduced a predicate, $\mathsf{SF}(a,s)$, to describe the result of performing the binary-valued sensing action $a$. $\mathsf{SF}(a,s)$ holds (returns *true*) iff the sensor associated with $a$ returns the sensing value 1 in situation $s$. Each sensing action senses some property of the domain. The property sensed by an action is associated with the action using a *guarded sensed fluent axiom* [5]. For example, the following two axioms
$\mathsf{InR1}(s) \to \big(\mathsf{SF}(\mathsf{SenseLight},s) \equiv \mathsf{Light1}(s)\big)$
$\neg\mathsf{InR1}(s) \to \big(\mathsf{SF}(\mathsf{SenseLight},s) \equiv \mathsf{Light2}(s)\big)$
can be used to specify that the $\mathsf{SenseLight}$ action senses whether the light is on in the room the agent is currently located in Example 1. Let $\phi[s]$ denote that fluent $\phi$ is believed at $s$, and $\mathsf{M}(a,s) = \mathsf{SF}(a,s) \equiv \phi[s]$ denote that the sensing result matches $s$, if $a$ is used to sense $\phi$.

Scherl and Levesque [19, 20] defined a successor state axiom for $B$, *an accessibility relation* on situations based on the possible-worlds semantics by [16], that shows how actions, including sensing actions, affect the beliefs of an agent.

$$B(s'', do(a,s)) \equiv \exists s', \qquad (1)$$
$$[B(s,s') \land s'' = do(a,s') \land (\mathsf{SF}(a,s') \equiv \mathsf{SF}(a,s))].$$

The situations $s''$ that are $B$-related to $do(a,s)$ are the ones resulting from doing action $a$ in a situation $s'$, s.t., the sensor associated with $a$ has the same value in $s'$ as it has in $s$, where (informally) a situation $s$ is B-related to another situation $s'$ if in situation $s$, the agent considers $s'$ is also possible. It follows the Kripke semantics.

Similar to [22], we take the following conventions about the *guarded action theories* $\Sigma$ consisting of: (A) successor state axioms for each fluent, and guarded sensed fluent axioms for each action; (B) unique names axioms for actions, and domain-independent foundational axioms; and (C) initial state axioms which describe the initial state of the domain and the initial beliefs of agents. A *domain-dependent fluent* means a fluent other than $B$ or $pl$ (the plausibility of situations that will be defined later), and a *domain-dependent formula* is one that only mentions domain-dependent fluents. We further assume that there is only one agent acting in a chosen domain, although the framework is capable of accommodating multiple agents. In addition, we assume that there are only finite initial situations considered by the agent.

## 3 Definitions and Axioms

In this section, we extend the framework in [22] to include a non-fixed plausibility operator to account for iterated belief changes in the situation calculus. The non-fixed plausibilities of situations enable us not to discard any situations, and hence the accessibility relation $B$ becomes very succinct. Surprisingly, it has a more expressive power than that in [22].

The revised accessibility relation $B$ is defined as follows.

$$B(s'', do(a,s)) \equiv \exists s', [B(s,s') \land s'' = do(a,s')]. \qquad (2)$$

It is clearly much simpler than Equation 1. Equation 2 ensures that if a set of situations are $B$-related, then their successors are also $B$-related, and so on. That is, if initially we have a set of $B$-related situations, then after the first action is taken, the successors of these situations are also $B$-related; after the second action is taken, the 2nd set of successors are again $B$-related.

We define a plausibility function $pl$ for each situation $s$ to measure how plausible an agent considers $s$ to be. The $pl$ function is in line with Spohn's ordinal conditional functions whose range is the set of natural numbers including 0. A lower $pl(s)$ value indicates a higher plausibility level of $s$. The $pl$ functions for the initial situations are given, with at least one $s$ having $pl(s) = 0$ showing the greatest plausibility level, using an initial state axiom as:

**Axiom 1** *(Initial State Axiom)* $\exists s, \mathsf{Init}(s) \land pl(s) = 0$.

Let $m \stackrel{def}{=} max_{s:\mathsf{Init}(s)}pl(s)$ be the maximum $pl$ value (hence the minimal plausibility level) of the initial situations and $t_{a,s} \stackrel{def}{=} min_{s':B(s',s)\land \mathsf{M}(a,s')=true}pl(s')$ be the minimum $pl$ value among the $B$-related situations of $s$ that match the sensing result of $a$. Plausibility values of successor situations are defined by the following successor state axiom:

**Axiom 2** *(Successor State Axiom)*
$$pl\big(do(a,s)\big) = \begin{cases} pl(s) - t_{a,s} & a \in \mathsf{SA} \land \mathsf{M}(a,s) = true \\ pl(s) + m + 1 & a \in \mathsf{SA} \land \mathsf{M}(a,s) = false \\ pl(s) & a \notin \mathsf{SA}, \end{cases}$$

Here and subsequently, we use $\mathsf{SA}$ to denote the set of all sensing actions, hence $a \in \mathsf{SA}$ indicates that $a$ is a sensing action. The successor state axiom for $pl$ says that if $a$ is not a sensing action, then the successor situation has the same plausibility as its predecessor; if $a$ is a sensing action and the current situation matches the sensing result, then the plausibility level of the successor situation should increase

(i.e., the plausibility value decreases), and the plausibility of the successor of the most plausible situation is changed to 0 (for normalization); if $a$ is a sensing action but the current situation does not match the sensing result, then the plausibility level of its successor should decrease, hence the plausibility value is increased by $m + 1$ to make sure it be greater than any of $B$-related situations that match the sensing result. This axiom follows a similar manner to OCF conditionalization [24].

This successor state axiom for $pl$ follows the spirit of the intuition stated for the plausibility settings in [22] that *if the accessible situation agrees with the actual situation upon the result of a sensing action, the plausibility of the accessible should increase (i.e., its $\kappa$-ranking should decrease), otherwise the plausibility should decrease*. However, this intuition was not implemented in the $pl$ function in [22] because *it is considered in conflict with positive and negative introspection of beliefs* [22]. Instead, the framework adopted a fixed plausibility value for each situation and its successors (although discarding some situations makes the beliefs changeable). Our Axiom 2 can be seen as an improvement over this fixed value framework and offers an opportunity to allow flexible changes of plausibilities of situations and hence beliefs (and still be able to handle introspections).

The belief operator $Bel$ is defined as follows.

**Definition 1**
$$Bel(\phi, s) \stackrel{def}{=} \forall s' \left( B(s', s) \land pl(s') = 0 \to \phi[s'] \right).$$

That is, $\phi$ is believed at $s$ when it holds at all the most plausible situations that are $B$-related to $s$. This is intuitively the same as the one defined in [22]. The only difference is that in our framework, the plausibility values are not fixed and we always have $B$-related situations which have a plausibility value 0.

For initializing the $B$-related situations, we have

**Axiom 3** $\forall s, s', \mathsf{Init}(s) \land \mathsf{Init}(s') \to B(s', s).$

That is, all initial situations are $B$-related to each other. Note that this axiom and the definition of $B$ relations together ensure that an initial situation will not be accessible by a non-initial situation.

Coupling with Equation 2, we have

**Theorem 1** $\forall s, s', \quad B(s', s) \to \left( \forall s'', \quad B(s'', s') \equiv B(s'', s) \right).$

That is, $B$ is transitive and Euclidean, hence the positive and negative introspection of beliefs holds in our framework.

Actually, Equation (2) and Axiom 3 imply that all situations having the same number of antecedents are mutually accessible. This means that the $B$-relation does not play a crucial role in this paper. But we keep this relation to make the analog with related work obvious.

## 4 Properties

In this section, we first present the key contribution of the paper, *noisy sensing tolerance*, which is beyond previous approaches. We then provide the counterparts of properties given in [22] which demonstrate that our framework faithfully extend their framework to noisy sensing situations. To make the comparisons easier to follow, in the corresponding subsections below, we have adopted many notations and results proposed in [22].

### 4.1 Noisy Sensing Tolerance

In this subsection, we show that our framework can deal with noisy sensing properly. More precisely, we prove that in the sense of probability, rare noisy sensing actions do not prevent an agent from deriving the actual situation.

To differentiate accurate sensing from noisy ones, for any sensing action $a$, let $a^T$ and $a^F$ denote that $a$ is accurate or noisy, respectively[2]. Let $P$ be a probability function that measures the statistical outcome of the accuracy of sensing actions performed by an agent [2]. For instance, $P(a = a^T) \approx 1$ is interpreted as *in a large number of experimental runs, the chance of $a$ being accurate is almost certain* and $P(a = a^F) \approx 0$ means it is very rare that $a$ returns a false result.

**Definition 2** *A situation calculus about a domain is called* sensing-sensitive *if there is a sequence (multi-set) of actions* $\mathsf{SEQ} = \{a_1, \cdots, a_n\}$ *such that:*

- $\forall a_i \in \mathsf{SEQ}$, *if* $a_i \in \mathsf{SA}$, *then* $a_i = a_i^T$.

- *Let* $S^*$ *be the actual situation[3], and $S$ be any situation, then $S \equiv S^*$ iff* $\mathsf{SF}(a_i, do(a_{i-1}, do(a_{i-2}, \cdots, S) \cdots)) = \mathsf{SF}(a_i, do(a_{i-1}, do(a_{i-2}, \cdots, S^*) \cdots))$, $\forall a_i \in \mathsf{SA} \cap \mathsf{SEQ}.$

Sensing-sensitive means that the actual situation can be uniquely determined by an agent after taking a sequence of accurate sensing actions and other physically executable actions. This issue was not explicitly considered in [22] but could be seen as a default assumption for their framework.

---

[2] Again here the phrase *accurate* or *noisy* is for the sensing device. The agent does not know whether each sensing is accurate or noisy, and he does not need to know. All he needs is the ability to train the device to obtain its probability of accuracy.

[3] The actual situation is the true state of the current environment. Generally an agent does not know the actual situation, but it is able to know given this sensing-sensitive property.

**Example 2** *(Example 1 Cont') Suppose all sensing actions are accurate, then the actual situation could be detected by a sequence of actions* SEQ = {SR, SL, LEAVE, SL} *where* SR *indicates that the agent senses whether it is in Room 1;* LEAVE *indicates that the agent leaves the current room for the other room; and* SL *is the same as in Example 1.*

*If* SR, SL, SL *are all accurate, then it is easy to see that* $S_2$ *satisfies that for any sensing action* $a_i$ *in* SEQ *(i.e.,* SR, SL, SL*), s.t.,*
$\mathsf{SF}(a_i, do(a_{i-1}, do(a_{i-2}, \cdots, S_2) \cdots)) = \mathsf{SF}(a_i, do(a_{i-1}, do(a_{i-2}, \cdots, S_0) \cdots))$, *while* $S_1$ *does not.*

Let $s^n$ be a situation believed by an agent after $n$ actions, and $S_0^n$ be the actual situation at the time these $n$ actions were taken. Let $I(X)$ be an *indication function* s.t. $I(X) = 1$ if $\Sigma \models X$ and $I(X) = 0$ otherwise. Now we show that the agent is able to detect the actual situation in the long run.

**Theorem 2** *If* $P(a = a^T | a \in \mathsf{SA}) \approx 1$ *and if the framework is sensing-sensitive, then* $\forall \phi, s, \lim_{n \to +\infty} \frac{\sum_{i=1}^{n} I(Bel(\phi, s^n) \equiv Bel(\phi, S_0^n))}{n} = 1$.

That is, if the noisy sensing actions are rare, and the framework is able to detect the actual situation by a sequence of accurate sensing actions, then after taking a finite number of actions, it can be sure that the actual situation can be detected in a probabilistic manner.

The proof is not difficult when the agent executes actions SEQ = $\{a_1, \cdots, a_k\}$ repeatedly. Indeed for any sensing action $a$, let $P(a = a^T | a \in \mathsf{SA}) > 1 - \epsilon$ where $\epsilon > 0$ is an arbitrarily small real number, then it is easy to show that the expected accuracy rate for question *whether the situation the agent detected is the actual situation*, is no less than $1 - k\epsilon$. Therefore, from a mathematical result on stochastic process, when $n \to +\infty$, $\frac{\sum_{i=1}^{n} I(Bel(\phi, s^n) \equiv Bel(\phi, S_0^n))}{n}$ reduces to the above expected rate and from $\epsilon \to 0$, we get the result.

As a corollary, if there is only a limited number of noisy sensing actions, then the agent is supposed to be able to detect the actual situation too. Let $|M|$ denote the cardinality of a set $M$.

**Corollary 1** *If* $|\{a : a \in \mathsf{SA} \wedge a = a^F\}| < +\infty$ *and the framework is sensing-sensitive, then* $\exists N > 0$, $\forall n > N$, $s^n \equiv S_0^n$.

However, if the probability of noisy sensing actions cannot be ignored, we have the following result.

**Theorem 3** *If the framework is sensing-sensitive, then* $\forall \phi, s, \lim_{n \to +\infty} \frac{\sum_{i=1}^{n} I(Bel(\phi, s^n) \equiv Bel(\phi, S_0^n))}{n} \geq \prod_{a_i \in \mathsf{SEQ}} P(a_i = a_i^T)$.

That is, even if the noisy sensing actions appear frequently, we may still have a (relatively good) chance to obtain the true situation in the long run.

### 4.2 Recovering the Beliefs in (Shapiro et al. 2011)

Similar to [22], for clarity, if there is no confusion, the situation argument of a fluent is often omitted in a belief operator, e.g., $Bel(\mathsf{InR1}, s)$. For comparison, let $\Sigma_S$ denote a guarded action theory used in [22] (with different definitions of $B$ relations and successive state axioms for $pl$) in contrast to using $\Sigma$ for a guarded action theory in our framework. Similarly, we use $pl_S$ and $Bel_S$ for plausibility functions and beliefs in [22].

In [22], the successor state axiom for $pl$ is defined as:

$$pl_S(do(a,s)) = pl_S(s). \qquad (3)$$

Its belief operator is defined as:
$Bel_S(\phi, s) \stackrel{def}{=} \forall s'$,
$B(s', s) \wedge (\forall s'', B(s'', s) \to pl_S(s') \leq pl_S(s'')) \to \phi[s']$.

Two axioms for initializing the $B$-related situations are used [22] to complete its framework.

**Axiom 4** *[22]* $\mathsf{Init}(s) \wedge B(s', s) \to (\forall s'', B(s'', s') \equiv B(s'', s))$.

This axiom requires that $B$-relations to be transitive and Euclidean initially. The aim is to obtain positive and negative introspection of beliefs.

**Axiom 5** *[22]* $\mathsf{Init}(s) \wedge B(s', s) \to \mathsf{Init}(s')$.

This axiom says that an agent considers nothing happened initially, hence any situations that are $B$-related to an initial situation are themselves initial.

Now we have the following result.

**Theorem 4** *If* $\forall a \in \mathsf{SA}$, $a = a^T$, *then for any* $\phi, s$, *if* $\Sigma_S \models Bel_S(\phi, s)$, *then* $\Sigma \models Bel(\phi, s)$.

This theorem shows that when all sensing actions are accurate, our framework truly discovers the beliefs which can be induced by the framework in [22].

The proof of the above theorem is not difficult. In fact, it is easy to show that if all sensing actions are accurate, then the situations that do not match the sensing results will henceforth have no chance to influence the change of beliefs, just like being discarded by the $B$-relations as done in [22].

### 4.3 Belief Revision

Belief revision studies how an agent's beliefs can be changed based on some new information if the new information must be believed. Sensing actions are a way of obtaining certain new information that an agent can use to revise its beliefs about the actual situation without actually changing the environment. Sensing actions do not change any fluents, instead they modify the $pl$ values corresponding to the degrees of beliefs of an agent. Therefore studying belief revision in situation calculus is a natural course for managing an agent's beliefs. In the following, we assume that for each formula $\phi$ to be revised, there is a corresponding sensing action.

**Definition 3** *(Uniform formula, adapted from [22]) A formula is* uniform *if it contains no unbound variables.*

**Definition 4** *(Revision action for $\phi$, adapted from [22]) A revision action $A$ for a uniform formula $\phi$ with respect to action theory $\Sigma$ is a sensing action that satisfies the following condition for every domain-dependent fluent $F$: $\Sigma \models \big[\forall s, SF(A,s) \equiv \phi[s]\big] \wedge \big[\forall s \forall \overrightarrow{x}, F(\overrightarrow{x},s) \equiv F(\overrightarrow{x}, do(A,s))\big]$, $\overrightarrow{x}$ is the set of arguments of $F$.*

It means that $A$ is a sensing action for $\phi$ which does not change any physical fluent. The following two theorems provided in [22] also hold in our framework.

**Theorem 5** *Let $\phi$ be a domain-dependent, uniform formula, and $A$ be a revision action for $\phi$ w.r.t. $\Sigma$, then we have:*
$\Sigma \models \big[\forall s, \phi[s] \rightarrow Bel(\phi, do(A,s))\big]$
$\wedge \big[\forall s, \neg\phi[s] \rightarrow Bel(\neg\phi, do(A,s))\big].$

This theorem proves that revision in our framework is handled adequately. That is, if the sensor indicates that $\phi$ holds, then the agent will believe that $\phi$ holds after performing $A$. Conversely, if the sensor shows that $\phi$ does not hold, then the agent will believe $\neg\phi$ after performing $A$. This theorem is also consistent with the framework in [19, 20].

**Theorem 6** *Let $A$ be a revision action for domain-dependent, uniform formula $\phi$ w.r.t. $\Sigma$, then the following sentence is satisfiable:*
$\Sigma \cup \{Bel(\neg\phi, S_0), Bel(\phi, do(A, S_0)), \neg Bel(FALSE, do(A, S_0))\}.$

This theorem shows that even if the agent believes $\neg\phi$ in $S_0$, it will believe $\phi$ after performing $A$ when action $A$ senses that $\phi$ is true, and still maintains consistent beliefs ($\neg Bel(FALSE, do(A, S_0))$).

### 4.4 Introspection

Like [19, 20, 22], our framework supports belief introspection.

**Theorem 7** $\Sigma \models \big[Bel(\phi, s) \rightarrow Bel(Bel(\phi), s)\big] \wedge \big[\neg Bel(\phi, s) \rightarrow Bel(\neg Bel(\phi), s)\big].$

This is not surprising, since in our framework the $B$-relation is transitive and Euclidean.

Note that in [22], the beliefs are induced from the most plausible $B$-related situations instead of from all $B$-related situations. The proof of the above theorem is simply similar to the proof of Theorem 26 in [22].

In [22], it is argued that variations of their formalization where plausibility values are updated lead to problems with introspection. They may produce counterintuitive results about future beliefs (cf. [22] for details). Note that variations mentioned in [22] have plausibilities between $B$-accessible situations, while our plausibility is assigned to single situations, so our framework does not have such weakness, as can be seen from Theorem 4 (showing that we will not produce counterintuitive results since the framework [22] does not) and Theorem 7 (showing that introspection holds).

### 4.5 Awareness of errors

As in [22], suppose that an agent believes $\neg\phi$ in $s$, however after performing a revision action $A$ in $s$, the agent discovers that $\phi$ is true and believes $\phi$. Then in $do(A,s)$, the agent should believe that in the previous situation $s$, $\phi$ was true, but it believed $\phi$ was false. In other words, the agent should realize that it was mistaken about $\phi$ in $s$.

**Definition 5** *([22])* $Prev(\phi, s) \stackrel{def}{=} \exists a, s',\ s.t.,\ s = do(a, s') \wedge \phi[s'].$

$Prev(\phi, s)$ denotes that $\phi$ was held in the situation immediately before $s$.

The following theorem provided in [22] also holds here.

**Theorem 8** *Let $A$ be a revision action for a domain-dependent, uniform formula $\phi$ w.r.t. $\Sigma$, then: $\Sigma \models \forall s, Bel(\neg\phi, s) \wedge Bel(\phi, do(A,s)) \rightarrow Bel(Prev(\phi \wedge Bel(\neg\phi)), do(A,s)).$*

The ability of belief update [22] is also provable from our framework. We omitted it here due to space limitation.

## 5 Example

We now extend Example 1 to illustrate our framework.

**Example 3** *Assumption: one agent with three actions: the agent leaves the current room and enters the other room (*LEAVE*); the agent senses whether it is in Room 1 (*SR*); the agent senses whether the light is on in the room it is currently located (*SL*). And for noisy sensing, we annotate it as* NSL. *Note that the agent itself does not know whether the sensing is accurate or not.*

*The successor state axioms and guarded sensed fluent axioms for the example are as follows:*
Light1$\big(do(a,s)\big) \equiv$ Light1$(s)$

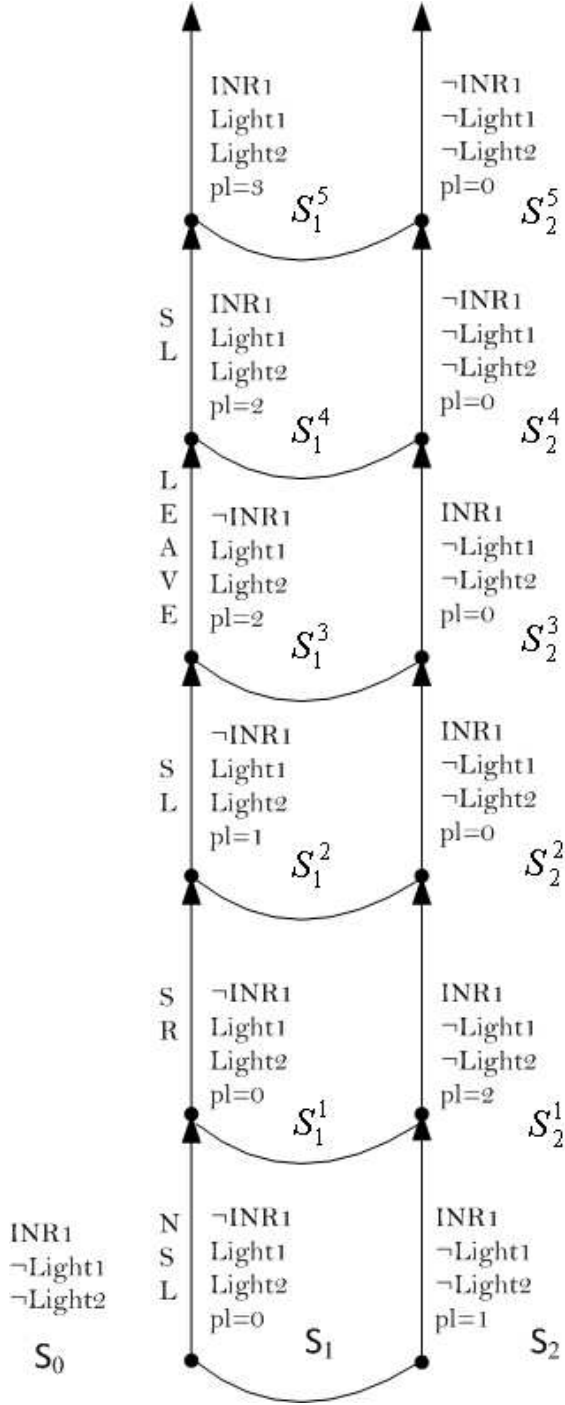

Figure 2: Extension of Example 1

$\mathsf{Light2}\big(do(a,s)\big) \equiv \mathsf{Light2}(s)$
$\mathsf{InR1}\big(do(a,s)\big) \equiv$
$\big((\neg\mathsf{InR1}(s) \wedge a = \mathsf{Leave}) \vee (\mathsf{InR1}(s) \wedge a \neq \mathsf{Leave})\big)$
$\mathsf{TRUE} \rightarrow \big(\mathsf{SF}(\mathsf{LEAVE},s) \equiv \mathsf{TRUE}\big)$
$\mathsf{InR1}(s) \rightarrow \big(\mathsf{SF}(\mathsf{SL},s) \equiv \mathsf{Light1}(s)\big)$
$\neg\mathsf{InR1}(s) \rightarrow \big(\mathsf{SF}(\mathsf{SL},s) \equiv \mathsf{Light2}(s)\big)$
$\mathsf{TRUE} \rightarrow \big(\mathsf{SF}(\mathsf{SR},s) \equiv \mathsf{InR1}(s)\big)$

*The initial states including the actual initial state are stated in Example 1. The initial state axioms for $S_1, S_2$ are:*
$\big(\exists s, Init(s) \wedge pl(s) = 0\big) \wedge \big(\forall s, Init(s) \wedge pl(s) = 0 \rightarrow \neg\mathsf{InR1}(s) \wedge \mathsf{Light1}(s) \wedge \mathsf{Light2}(s)\big)$
$\big(\exists s, Init(s) \wedge pl(s) = 1\big) \wedge \big(\forall s, Init(s) \wedge pl(s) = 1 \rightarrow \mathsf{InR1}(s) \wedge \neg\mathsf{Light1}(s) \wedge \neg\mathsf{Light2}(s)\big)$

*State axioms for other states are similar and omitted here. For simplicity, we only give an informal explanation of the first two steps in the process depicted by Figure 2.*

- *The first action is a noisy light sensing, which gives the result that* the light is on in the room the agent is located. *As the agent thinks it is in Room 2, $S_1$ matches the sensing result whilst $S_2$ does not. Based on Axiom 2 (Sec. 3), the plausibility value of $S_2^1$ increases to 2.*

- *The second action is the room sensing, which tells the agent that it is in Room 1 not Room 2. Hence the agent reassigns $S_2^2$ with plausibility 0 and $S_1^2$ with 1. Now its belief is in accord with $S_2^2$.*

*In fact, we can see that the action sequence $\mathsf{SR}, \mathsf{SL}, \mathsf{LEAVE}, \mathsf{SL}$ makes the situation calculus sensing sensitive. Also note that the framework in [22] cannot proceed after the sensing action $\mathsf{SR}$.*

## 6 Related Work

In [3], the problem of noisy sensors is also studied and a probabilistic method is applied for such situations. That is, the probability of a sensing action result follows a Gaussian distribution. In addition, the beliefs induced from the situations are also probabilistic. This approach is extended in [23] to study its properties and allows for using conditional probability densities in the noisy sensor readings. To some extent it could be seen as a probabilistic counterpart of our framework although noisy sensing tolerance is not addressed in these papers.

In [21], the noisy sensor problem is also studied by adding plausibility values into the $B$-relation, i.e., $B(s', n, s)$ which means that in $s$, the agent thinks $s'$ is possible with a plausibility value $n$. Similar to our approach, noisy sensing results will affect the $n$ value based on whether the situation matches the sensing result. However, in [22], it is discussed that this kind of accessibility relations, together with the transitive and Euclidean condition required by belief introspection, conflicts with any reasonable plausibility update scheme for accurate sensors.

In [6], a fluent calculus framework is proposed to deal with the problem of observations contradicting the model which is to some extent similar to noisy sensing. However, in the formalism, the state ranking change axioms need the details of actions which make the formalism somehow lose generality.

# 7 Conclusion

In this paper, we proposed a framework which can deal with noisy sensing actions and can handle nested beliefs, belief introspection, mistaken beliefs, belief revision and belief update. In addition, we show that rare noisy sensing does not prevent an agent from detecting the actual situation, and limited noisy sensing allows for a certain degree of detecting the actual situation. Moreover, we show that our framework induces what can be discovered by the framework of [22] when all sensing actions are accurate.

Due to space limitation, in this paper, we do not provide the comparison of our framework with the various belief change postulates, e.g., AGM belief revision postulates [1], KM belief update postulates [7], DP iterated belief revision postulates [4], epistemic state revision postulates [12], belief change postulates [10, 11], etc. However, we can prove that our framework satisfies most of the postulates mentioned above. Especially, DP's C2 postulate is satisfied in our framework, whilst in [22], C2 cannot be defined.

For future work, we want to extend our framework to multiple agent scenarios. It is also very interesting to study whether an agent can realize that the previous sensing is inaccurate in belief introspection. In addition, a study of the relationship between our framework and the Partially Observable Markov Decision Process (POMDP) [18] could be helpful.